\documentclass[runningheads]{llncs}
\usepackage{xcolor}
\usepackage[numbers]{natbib}
\usepackage{graphicx}
\usepackage{tabularx}
\usepackage{multirow}
\usepackage{amsmath}
\usepackage{algorithm}
\usepackage{algpseudocode}
\newtheorem{defn}{Definition}
\usepackage{hyperref} 
\begin{document}
\title{Explainability of Highly Associated Fuzzy Churn Patterns in Binary Classification}
\titlerunning{Explainability of Highly Associated Fuzzy Churn Patterns}
%
\author{D.Y.C. Wang\inst{1}\thanks{This paper is an extended version of a work originally presented at the 6th International Workshop on Utility-Driven Mining and Learning (held in conjunction with the 28th Pacific-Asia Conference on Knowledge Discovery and Data Mining - PAKDD 2024) on May 7, 2024.} \and Lars Arne Jordanger\inst{1} \and
Jerry Chun-Wei Lin\inst{1}}
\authorrunning{Wang et al.}
%
\institute{Western Norway University of Applied Sciences, Bergen, Norway\\ 
\email{ycw@hvl.no; Lars.Arne.Jordanger@hvl.no; jerrylin@ieee.org}\\}
%
%
\maketitle              
\begin{abstract}
Customer churn, particularly in the telecommunications sector, influences both costs and profits. As the explainability of models becomes increasingly important, this study emphasizes not only the explainability of customer churn through machine learning models, but also the importance of identifying multivariate patterns and setting soft bounds for intuitive interpretation. The main objective is to use a machine learning model and fuzzy-set theory with top-\textit{k} HUIM to identify highly associated patterns of customer churn with intuitive identification, referred to as Highly Associated Fuzzy Churn Patterns (HAFCP). Moreover, this method aids in uncovering association rules among multiple features across low, medium, and high distributions. Such discoveries are instrumental in enhancing the explainability of findings. Experiments show that when the top-5 HAFCPs are included in five datasets, a mixture of performance results is observed, with some showing notable improvements. It becomes clear that high importance features enhance explanatory power through their distribution and patterns associated with other features. As a result, the study introduces an innovative approach that improves the explainability and effectiveness of customer churn prediction models.

\end{abstract}
\section{Introduction}\label{sec:intro}
Customer churn refers to consumers switching services, often due to dissatisfaction, better options or personal preference \cite{AMIN2023110103}. Fluctuations in churn rates can have a major impact on costs and profits. An example from the telecom sector: while churn rates are typically between $20\%$ and $40\%$, it costs more than $5$ - $6$ times more to retain customers, and a mere $5\%$ reduction can lead to a $25\%$ increase in profits \cite{AMIN2023110103}. 
Customer churn prediction (CCP) is mostly a binary classification problem, unless there is a detailed breakdown of the reasons for customer churn into multiple labels, in which case it can be handled as a multi-class classification problem. Howerver, CCP not only determines whether a customer is likely to churn and identifies the reasons, but also helps in implementing customer retention strategies to mitigate the economic loss. Nowadays, it is more important to understand how the models explain the results of churn prediction.

For the interpretation of CCP, it is essential to understand the complexity of the numerous influencing variables. These variables not only function independently of each other, but also exhibit interrelated patterns between different distributions of characteristics. However, relying only on the explainability of machine learning (ML) models may not fully uncover these nuanced associations, especially in the context of association rule mining. Furthermore, predictive models work primarily with numerical variables that lack definitive boundaries for intuitive interpretation, which presents a particular challenge. For example, the lack of a clear threshold for categorizing a customer's call duration as high or low leads to ambiguity. Accordingly, there are two main needs in this study beyond pure prediction and explainability. First, to classify data by defining soft boundaries rather than using traditional unambiguous thresholds to provide a clearer and more actionable perspective for stakeholders, and second, to identify patterns that reveal underlying associations.

\begin{figure*}[!ht]
		\centering
		\centerline{\includegraphics[scale=0.8]{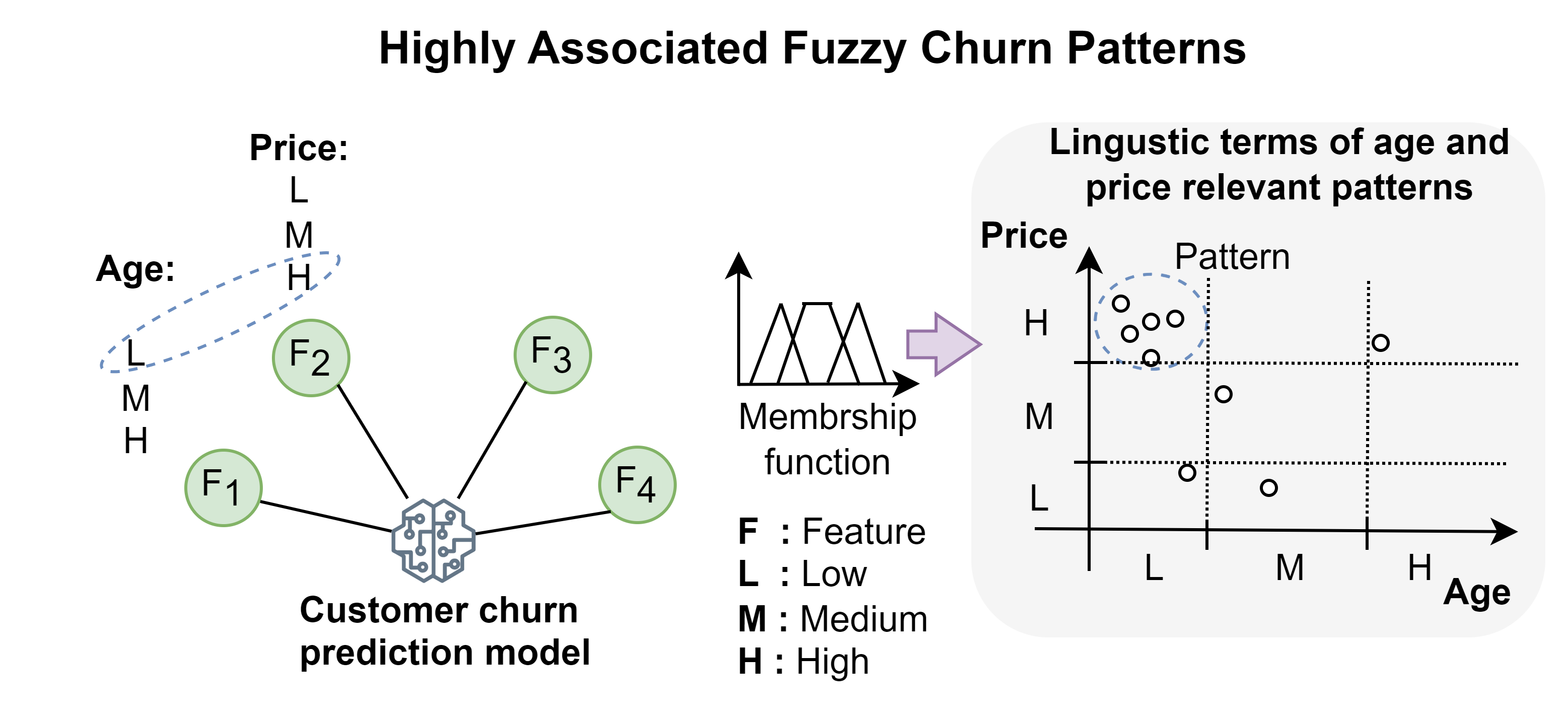}}
		\caption{An illustrative example of highly associated fuzzy churn patterns. The figure simply illustrates the concept of HAFCP. In the diagram, 'L', 'M', and 'H' represent low, medium, and high respectively, which are calculated by the membership function in fuzzy-set theory.}
		\label{fig1}
	\end{figure*}

The main goal is to develop an integrative framework that connects ML models with top-\textit{k} high utility itemset mining (HUIM). In a previous study, a churn prediction pattern (CPP) \cite{wang2023} was defined as a set of features that can influence several definable indicators for customer churn evaluation. This study, the extracted patterns are defined as Highly Associated Fuzzy Churn Patterns (HAFCP), which are conceptually illustrated in Fig. \ref{fig1}. This aims not only to provide revealing insights into the underlying data, but also to increase the predictive capability of the model. A study using a similar concept of HUIM and statistical learning has already been proposed in the field of inductive logic programming (ILP) \cite{shakerin2019induction}, but while their focus was primarily on rule discovery and they used quite different datasets than this study, our intention is to improve the explainability and strengthen the predictive capabilities of the model.

\section{Related Work}
\subsection{Explainability of SHAP Value}
The latest developments \cite{saeed2023explainable} in eXplainable AI (XAI) are gradually making ML and artificial intelligence (AI) transparent. In particular, model-agnostic methods such as SHapley Additive exPlanations (SHAP) facilitate sophisticated modeling, understanding and representation of complex events and systems \cite{LI2022101845}. A unified approach that uses SHAP and feature differences to explain ML predictions based on game theory \cite{lundberg2017unified}. The overall importance of a feature in a model can be calculated by taking the mean of the absolute SHAP values for that feature in all instances. This gives a general indication of the importance of a particular feature in terms of its impact on the output of the model. In recent studies \cite{guliyev2021customer, peng2022research, sina2022model}, various cases have been observed where SHAP values were used to explain customer churn prediction models. Peng and Peng \cite{peng2022research} integrated GA-XGBoost and SHAP to solve the  problems faced by customer churn in telecom business management. Mirabdolbaghi  and Amiri \cite{sina2022model} presented how the features affect the model’s output by the SHAP method. Guliyev and Tato{\u{g}}lu \cite{guliyev2021customer} considered customer churn prediction using banking data, and applied SHAP values to support model's interpretability.

\subsection{Fuzzy-Set Theory and Fuzzy Pattern Mining}
In contrast to traditional classification, where data is categorized into specific sets, fuzzy-set theory assigns data based on membership functions that determine the degree to which each piece of data belongs to a particular set \cite{jang1997neuro}, \cite{zadeh1978fuzzy}. Common methods include the Gaussian fuzzy function and the triangular fuzzy function for classification. For data that follows a normal distribution, the Gaussian fuzzy function is usually preferred \cite{systems10060258}. Leveraging the principles of fuzzy set theory \cite{hong2003fuzzy} in the context of utility mining enables the development of a data-driven model that is not only intuitive for human understanding but improves efficiency in mining operations. Wang et al. first attempt to discover high utility patterns from quantitative database \cite{5277408}. Then, Gan et al. proposed an explainable fuzzy utility mining (FUM) for fuzzy high-utility sequence patterns \cite{gan2021explainable}.

\subsection{Top-\textit{K} High Utility Itemset Mining}
High Utility Itemset Mining (HUIM) is an essential topic in data mining that aims to discover itemsets that bring high profit or utility in transactional databases \cite{fournier2019survey}. Traditional HUIM focuses on finding all itemsets that exceed a user-defined utility threshold. However, setting an appropriate utility threshold is challenging, which led to the emergence of the top-\textit{k} model for mining high utility itemsets \cite{tseng2015efficient}. Traditional utility mining can be overwhelming due to the generation of numerous itemsets, while the top-\textit{k} approach selectively highlights the most relevant k-itemsets \cite{liu2012mining}. By focusing on high utility itemsets using the top-\textit{k} model, users can gain insights into highly profitable or important items and thus improve their decision-making processes accordingly.

\section{The Developed HAFCP Model}
In this section, this study describes the analytical framework for identifying strongly associated churn patterns with fuzzy logic using predictive feature importance, fuzzy-set theory, and top-\textit{k} high-utility itemset mining.
\subsection{Problem Statement}
This paper focuses exclusively on the binary classification problem for predictive modeling in the methodology section. Starting from a labeled customer churn dataset, this study aims to uncover associated patterns from categorical and numerical features by constructing a predictive model using fuzzy-set theory \cite{jang1997neuro} and leveraging top-\textit{k}HUIM \cite{liu2012mining}. To provide a more intuitive understanding and explanation of the data, we convert numerical features into categorical levels such as high, medium and low using fuzzy-set theory \cite{jang1997neuro}. This conversion facilitates the use of the HUIM approach by transforming the original numerical data into corresponding frequencies of low, medium and high linguistic terms, thereby improving the explainability and relevance of the patterns found. Furthermore, we intend to use the global feature importance — a measure that quantifies the overall contribution of each feature to the ML model performance. Then, the HAFCPs are identified using the proposed method as the final output. Here is the mathematical definition of HAFCP:

\begin{defn}(Highly Associated Fuzzy Churn Pattern (HAFCP))\\
Consider a set of features \( V \), where each numerical feature \( v_i \) $\in$ \( V \) can be mapped to a linguistic term such as Low (L), Medium (M), or High (H). Let \( B(v_i) \) represent the application of the maximum cardinality criterion \cite{giacometti2019mining} to the membership degree of the numerical feature \( v_i \). Given a utility function \( U \) and a pattern \( P \), with \( P \subseteq V \), we can rank all patterns based on their computed values. The top-\textit{k} high utility patterns are then generated. HAFCP is defined as:
\begin{equation}
\begin{aligned}
HAFCP(K) = \{ & P_1, P_2, \ldots, P_K \mid P_i = U^{(i)} \land \\
& \forall v_j \in P_i, B(v_j) \in \{L, M, H\} \},
\end{aligned}
\end{equation}
where \( U^{(i)} \) represents the \( i^{th} \) highest value pattern, \( P_1 \) is the top-1 high utility pattern, \( P_2 \) the top-2 high utility pattern, and so forth. The condition \( B(v_j) \in \{L, M, H\} \) ensures that variables in each pattern are distinctly categorized, e.g., as “Low”, “Medium”, or “High” based on the defined membership functions.
\end{defn}
Using a customer dataset with features like 'Age' and 'Price', we aim to identify patterns linked to customer churn. Fuzzy-set theory categorizes 'Age' and 'Price' into 'Low', 'Medium', and 'High', where 'Age' represents different age groups and 'Price' different spending levels. With HUIM, we uncover patterns like \{'Age\_L', 'Price\_H'\}, suggesting that younger customers making high-priced purchases are more likely to churn. The patterns help us understand specific customer behaviors at higher risk of churn, guiding more focused retention strategies.

\subsection{Analytical Framework for HAFCP Mining}
In this section, we explain the methodology for identifying HAFCP using fuzzy-set theory and feature score evaluation. This approach utilizes the insights and knowledge presented in the previous section to provide a comprehensive analytical framework. Fig. \ref{fig2} shows the comprehensive structure of the proposed framework and the links between each phase. The framework for mining HAFCP consists of four main steps. First, a churn prediction model is built to determine the importance of the characteristics. Then, a dataset based on fuzzy-set theory is transferred. Next, top-\textit{k} HUIM is applied \cite{tseng2015efficient}. This approach aims to identify patterns that are strongly associated with frequent items and having high feature importance as determined by a trained ML model. Algorithm \ref{algo1} provides a detailed explanation of how the developed HAFCP is processed. In the upcoming subsubsections, this methodology will be further elaborated upon, with additional explanations and a simple example provided to show the procedure of the designed HFACP.

\begin{algorithm}[!ht]
\caption{HAFCP Mining}\label{algo1}
\begin{algorithmic}[1]
\State \textbf{Input:} Dataset $D$ with customer features and churn labels, Machine learning model $M$, SHAP Explainer $E$
\State \textbf{Output:} Top-$k$ highly associated churn patterns $P$

\Procedure{Mine HFACPs}{$D$, $M$, $E$}
    \State $D \gets \text{Read dataset}$
    \State $C \gets \text{Factorize categorical features in } D$
    \State $X, y \gets \text{Split } D \text{ into features and target}$
    \State Initialize $M$
    \State Train $M$ on $(X, y)$
    \State $\hat{y} \gets \text{Predict churn using } M$
    \State $\text{SHAP values} \gets E(M, X)$
    \State $\bar{S} \gets \text{Calculate mean absolute SHAP values for } X$
    \State $T \gets \text{Form profit table from } \bar{S}$
    \For{each $x_i \in X$}
        \State $\text{normality} \gets \text{Test normality of } x_i$
        \State $\mu \gets 
            \begin{cases} 
                \mu_T & \text{if } \text{normality is false} \\
                \mu_G & \text{otherwise}
            \end{cases}$
        \State $x'_i \gets \text{Fuzzy transform } x_i \text{ using } \mu$
        \State $x''_i \gets \text{Assign } L, M, \text{ or } H \text{ to } x'_i \text{ based on max membership}$
        \State $\text{One-hot encode } x''_i$
    \EndFor
    \State $F \gets \text{Collect transformed features}$
    \State $F \gets F \cup C$
    \State Define HUIM algorithm $H$
    \State $P \gets H(F, T, k)$
    \State \textbf{return} $P$
\EndProcedure
\end{algorithmic}
\end{algorithm}
\subsubsection{A. Feature Importance Calculation}
\begin{figure*}[!ht]
		\centering
		\centerline{\includegraphics[scale=0.61]{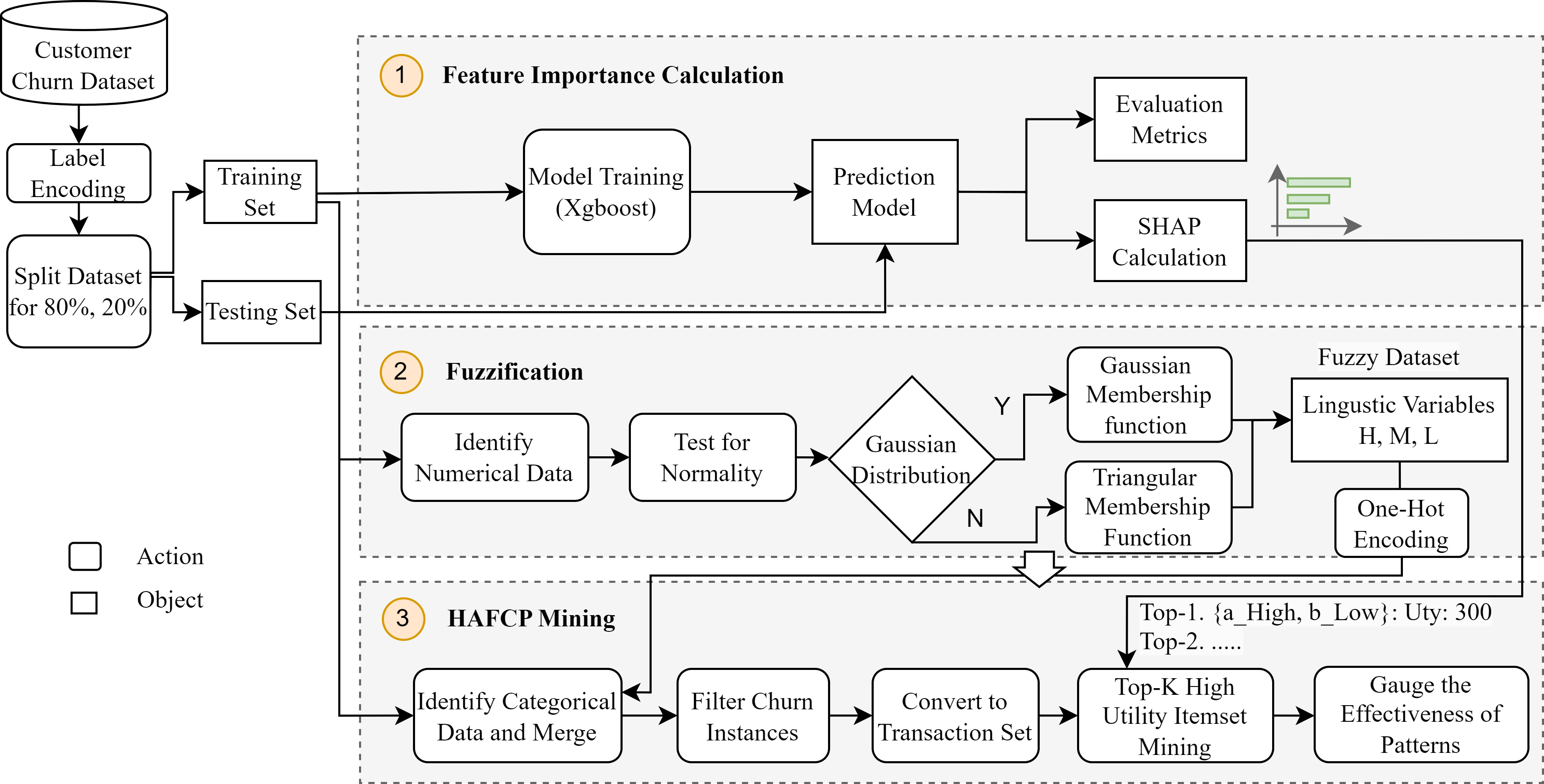}}
		\caption{The proposed framework for mining HAFCPs: This comprehensive framework outlines the process of extracting HAFCPs, detailing the methodology from initiation to conclusion. To facilitate understanding, the extension work will present a straightforward algorithmic example. This example will elaborate each step of the process, providing a clear explanation of how HAFCPs are identified and derived within our framework.}
		\label{fig2}
	\end{figure*}

The process begins with a specific dataset on customer churn with the label churn or do not churn. Once this dataset has been captured, encoding of the labels for potentially categorical data is required to convert it into a machine-readable format. After data conversion, the dataset is divided into two parts: $80\%$ is assigned to the training set, while the remaining $20\%$ is assigned to the testing set. In Fig. \ref{fig2} part $1$ outlines a series of steps for model construction and evaluation. The process begins with model training using the XGBoost \cite{chen2016xgboost}, a gradient boosting framework, to train the model using the specified training data. After training, the model is applied to the testing dataset and its performance is evaluated using various metrics such as accuracy, recall rate, precision, etc. At the same time, the importance of the features of the model is calculated using various techniques. Particular attention is paid to SHAP values, which are a standardized measure of feature importance. Other methods including the inherent information gain and the weight of the tree-based model are mentioned \cite{altmann2010permutation}. The compilation of these values illustrates the contribution of each feature to the prediction of the model under different evaluation methods. The revealed values of the features can be considered as the gain values in high-utility itemset mining since they represent a quantified importance in prediction and have been applied in the study \cite{shakerin2019induction}.

\subsubsection{B. Data Transformation}
The second phase is primarily concerned with converting numerical data columns into a more interpretable fuzzy dataset, which is then translated into lingustic terms based on fuzzy set theory. The process begins with the identification of the numerical data columns, which are subjected to a normality test called "Shapiro–Wilk test" \cite{shapiro1965analysis}. Depending on the result of the normality test, the data distributions are classified as either Gaussian or non-Gaussian. For data with a Gaussian distribution, a Gaussian fuzzy membership function is used \cite{jang1997neuro}. Conversely, a triangular fuzzy membership function \cite{jang1997neuro} is used for data with a non-Gaussian distribution due to simplicity and clearly defined range. Specifically, these numerical features are translated into three linguistic variables, including High, Medium and Low with their own memberships. The culmination of this phase is a fuzzy dataset that categorically represents the originally identified numerical data in the form of these linguistic variables. Subsequently, one-hot encoding \cite{hancock2020survey} is inevitably used to implement the data for mining. One-hot encoding is a process in which categorical data is converted into a binary matrix, with each category represented by a unique binary column. After one-hot encoding, the data columns can be left with defuzzified binary values. This study discusses triangular membership functions in fuzzy set theory, similar in concept to Gaussian membership functions, as referenced in \cite{jang1997neuro}. These functions are essential in determining the membership degree of elements in a fuzzy set. We define a fuzzy set A on a universe of discourse X. The triangular membership function of A, denoted \(\mu_A(x)\), uses three parameters $a$, $b$, and $c$ to represent the vertices of the triangle. It is defined as:

\begin{equation}
\mu_A(x) = 
\begin{cases} 
0 & \text{if } x \leq a \\
\frac{x - a}{b - a} & \text{if } a < x \leq b \\
\frac{c - x}{c - b} & \text{if } b < x < c \\
0 & \text{if } x \geq c
\end{cases}
\end{equation}

where $\frac{x - a}{b - a}$ represents the ascending side of the triangle (low to medium), $\frac{c - x}{c - b}$ the descending side (medium to high), and $0$ for values outside $[a, c]$. Using \(\mu_A(x)\), we categorize elements into low, medium, or high classes. The linguistic term for a given $x$ is determined by:

\begin{equation}
\text{Linguistic Term}(x) = 
\begin{cases} 
L & \text{if }\mu_A(x) \text{ is maximum in the low set} \\
M & \text{if }\mu_A(x) \text{ is maximum in the medium set} \\
H & \text{if }\mu_A(x) \text{ is maximum in the high set},
\end{cases}
\end{equation}
where \(\mu_A(x)\) represents a membership function used in this paper.
\subsubsection{C. Top-\textit{K} High Utility Itemset Mining}
In the third phase, the process focuses mainly on extracting meaningful patterns from the quantitative data. The process begins with the identification of original categorical features within the customer churn dataset. Following this identification, the categorical data is then merged with the previously created fuzzy dataset to create a consolidated dataset. Since our goal is to find churn patterns, we filter out the data that contains churn instances. These filtered data instances are then converted into a frequency itemset format, a standard format that is compatible with mining algorithms. At the end of this phase is the implementation of the top-\textit{k} HUIM \cite{tseng2015efficient} process. Specifically, the top-\textit{k} HUIM algorithm begins by calculating the utility of each itemset in the dataset. A single feature of the data belongs to one of the categories low, medium, or high, we will treat the quantitative value of the feature as 1; if not, it will be treated as 0. The unit of profit is determined using the corresponding feature importance. The utility of an itemset is usually defined as the sum of the products of the individual utilities of the items and their respective quantities in the transactions. This calculation reflects both the importance and the frequency of the items within the dataset. After calculating the utility values, the algorithm creates a ranking list of these item groups based on their utility values. From this ranking, the top-\textit{k} itemsets with the highest utility values are selected. From now on, we refer to the patterns identified using top-\textit{k} HUIM as HAFCPs.
\subsubsection{D. Pattern Evaluation and Model Enhancement}
In the last stage, before we move on to explaining the patterns obtained from the mining process, it is key to understand how we measure their meaning and effectiveness. This step is about understanding the key metrics to evaluate the quality and relevance of the identified patterns. With the utility values calculated, HAFCP can examine the identified patterns in more detail, particularly those associated with churn events. The aim is not only to identify which patterns occur frequently, but also to understand their relationship and significance in relation to churn events. Unlike traditional ML explainability tools, we can intuitively use labels such as low, medium and high, which are based on membership functions and allow us to clearly evaluate these combination patterns. Once we have understood and explored the patterns associated with churn, it is important to evaluate how these patterns and insights can improve the predictive model. To do this, we need to repeat the feature engineering phase of the model to integrate the insights from the patterns. Essentially, the aim here is to determine whether the inclusion of these patterns as features improves the performance of the model or provides a more refined understanding of churn events. In this study, we utilize the identified top-\textit{k} patterns to facilitate feature development. These patterns are converted into a new feature for use in ML models. This approach is also the reason why we exclusively use the training set for utility mining. In this study, we create a new feature that characterizes the presence of a particular HAFCP. In our experiments, we will observe whether these newly introduced features can effectively improve the model performance.

\subsubsection{E. A Simple Example}\label{subsec:simpleexample}
\begin{figure*}[!ht]
		\centering
		\centerline{\includegraphics[scale=0.4]{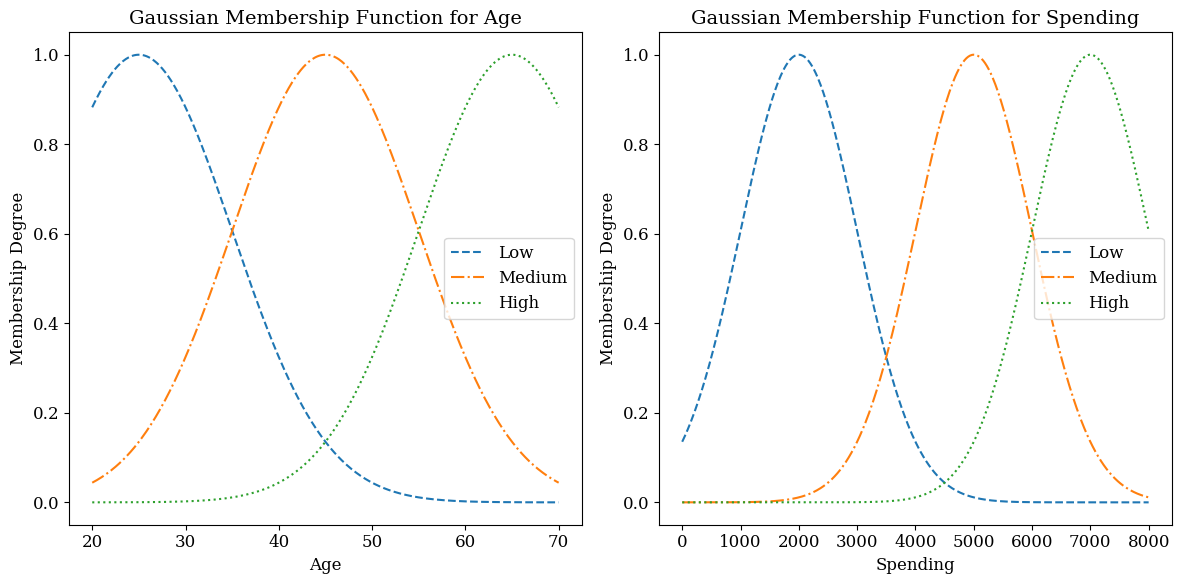}}
		\caption{An example of Gaussian membership functions representing linguistic terms (\textit{Low}, \textit{Medium}, and \textit{High}) for Age and Spending Variables in fuzzy-set theory}
		\label{fig5}
	\end{figure*}

In this subsection, we will demonstrate how to calculate and identify HAFCP using a simple example. Let us assume a customer churn training dataset containing the identities of the customers. For categorical variables, it contains the locations where customers shop ($South = S$, $Central = C$ and $North = N$). For numerical variables, it contains the age of the customer and their total spent. In addition, there is a label indicating whether the customer has churned, marked as $0=``Non-Churn''$ or $1=``Churn"$. A simple example is shown in Table \ref{tab:simpleExample}. Following the steps of the analysis framework from the previous subsection, a label encoding of the dataset must be carried out firstly (e.g., defining the shop location $N=1$, $S=2$, $C=3$). Then, we create a ML prediction model based on the XGBoost \cite{chen2016xgboost}. After training the ML model, we obtained the following feature importance: purchase location: $0.2$, age: $0.5$, purchase amount: $0.3$. The feature importance here is the value of profit used in high utility mining.
\begin{table}[ht]
    \caption{A simple customer churn dataset.}
    \centering
    \begin{tabular}{|c|c|c|c|c|}
    \hline
    \textbf{ID} & \textbf{Shop Location} & \textbf{Age} & \textbf{Spending} & \textbf{Churn (Given)} \\
    \hline
    A & N (1)& 25 & 5000 & 1 \\
    B & S (2)& 30 & 3000 & 0 \\
    C & N (1)& 28 & 4500 & 1 \\
    D & C (3)& 55 & 7000 & 0 \\
    E & N (1)& 60 & 1000 & 1 \\
    F & S (2)& 35 & 6500 & 0 \\
    G & N (1)& 40 & 5500 & 1 \\
    H & C (3)& 65 & 3500 & 0 \\
    I & S (2)& 23 & 4500 & 1 \\
    J & N (1)& 50 & 3000 & 1 \\
    \hline
    \end{tabular}
    \label{tab:simpleExample}
\end{table}
Next, it needs to categorize our numerical variables into easily interpretable linguistic terms such as low, medium and high using fuzzy set theory \cite{zadeh1978fuzzy} based on pre-defined membership functions. First, we need to confirm the distribution of the numerical data columns before we can decide whether to use a Gaussian membership function or a triangular membership function. For $``Age''$ and $``Spending''$ columns we first need to check if they follow a normal distribution. In this example, we used the Shapiro-Wilk test \cite{razali2011power}, which is suitable for small samples of less than $50$. The results show that both columns are consistent with the null hypothesis. Therefore, they are considered to be normally distributed. Subsequently, two normally distributed columns are divided into three different categories using the Gaussian fuzzy membership function \cite{jang1997neuro}: \textit{low}, \textit{medium} and \textit{high}, which enables a more differentiated analysis and interpretation. Fig. \ref{fig5} illustrates an example of Gaussian membership functions used to categorize the numerical variables $``Age''$ and $``Spending''$ into linguistic terms such as \textit{Low}, \textit{Medium}, and \textit{High} based on fuzzy set theory. The results according to the fuzzy-set theory and classification can be found in Table \ref{tab:simpleExamplefuzzy}.

\begin{table}[ht]
\caption{A simple customer churn dataset after transformation by the fuzzy-set theory.}
\centering
\begin{tabular}{|c|c|c|c|c|}
\hline
\textbf{ID} & \textbf{Shop Location} & \textbf{Age\_fuzzy} & \textbf{Spending\_fuzzy} & \textbf{Churn (Given)} \\
\hline
A & N (1) & Low  (0.97)& Medium (0.97) & 1 \\
B & S (2) & Low (0.99)& Low (0.99)& 0 \\
C & N (1) & Low (0.99)& Medium (1)& 1 \\
D & C (3) & Medium (0.99)& High (0.66)& 0 \\
E & N (1) & High (0.92)& Low (0.49)& 1 \\
F & S (2) & Medium (0.98)& High (0.82)& 0 \\
G & N (1) & Medium (0.98)& Medium (0.99)& 1 \\
H & C (3) & High (0.76)& Low (0.97)& 0 \\
I & S (2) & Low (0.93)& Medium (1)& 1 \\
J & N (1) & High (0.97)& Low (0.99)& 1 \\
\hline
\end{tabular}
\label{tab:simpleExamplefuzzy}
\end{table}
Since the problem we defined focuses on the utility patterns of churn, we first need to filter out data instances where $``Churn'' = 0$. As shown in Table \ref{tab:simpleExamplelabelencod} by simply performing One-hot encoding \cite{hancock2020survey}, we can convert our dataset into a binary dataset.

\begin{table}[ht]
\caption{Generation of frequent itemsets from customer churn data, where SL is labeled as shop location, L as low, M as medium and H as high.}
\centering
\small 
\begin{tabular}{|c|c|c|c|c|c|c|c|c|c|c|}
\hline
\textbf{ID} & \textbf{SL\_C} & \textbf{SL\_N} & \textbf{SL\_S} & \textbf{Age\_L} & \textbf{Age\_M} & \textbf{Age\_H} & \textbf{Spend\_L} & \textbf{Spend\_M} & \textbf{Spend\_H} \\
\hline
A & 0 & 1 & 0 & 1 (0.97) & 0 & 0 & 0 & 1 (0.97) & 0 \\
C & 0 & 1 & 0 & 1 (0.99) & 0 & 0 & 0 & 1 (1) & 0 \\
E & 0 & 1 & 0 & 0 & 0 & 1 (0.92)& 1 (0.49)& 0 & 0 \\
G & 0 & 1 & 0 & 0 & 1 (0.98)& 0 & 0 & 1 (0.99)& 0 \\
I & 0 & 0 & 1 & 1 (0.93)& 0 & 0 & 0 & 1 (1)& 0 \\
J & 0 & 1 & 0 & 0 & 0 & 1 (0.97) & 1 (0.99)& 0 & 0 \\
\hline
\end{tabular}
\label{tab:simpleExamplelabelencod}
\end{table}

With the transactions of the binary dataset, considered as itemsets, and the profit table, HUIM can be applied in this case to determine the associated fuzzy feature combinations with the maximum total utility. In contrast to setting a minimum threshold as in traditional HUIM methods, we use a top-\textit{k} \cite{tseng2015efficient} approach that allows us to identify the patterns with the highest utility in descending order by the pre-defined \textit{k} value. To save computational cost and memory usage, rows consisting exclusively of zeros can be eliminated directly, since multiplication by the utility always results in zero. Therefore, the columns $``SL\_M''$ and $``Spending\_H''$ should be omitted before we proceed with the mining calculations. In this case, the utility of a frequent itemset is calculated by summing the product of the presence of each item and the corresponding profit from a given profit table, and searching for the top-\textit{k} HUIs by iteratively expanding the itemsets and evaluating their utility. After each iteration, only the top-\textit{k} itemsets with the highest utility are retained for further expansion. This process is repeated until no more itemsets can be expanded. Table \ref{tab:simpleExampleresult} shows the results for this example with slight variations noted between the two result sets for the utility value of the top-3 patterns.

\begin{table}[ht]
    \caption{Top-5 highly associated fuzzy churn patterns using binary value.}
    \centering
    \begin{tabular}{|c|c|c|}
    \hline
        \textbf{Rank} & \textbf{Highly Associated Fuzzy Churn Patterns} & \textbf{Utility} \\
        \hline
        1 & \{``Age\_Low", ``Spending\_Medium"\} & 2.0 \\
        2 & \{``Age\_Low", ``SL\_N", ``Spending\_Medium"\} & 2.0 \\
        3 & \{``SL\_N", ``Spending\_Medium"\} & 2.0 \\
        4 & \{``Age\_Low", ``SL\_N"\} & 1.0 \\
        5 & \{``Age\_Medium", ``SL\_N"\} & 1.0 \\
        \hline
    \end{tabular}
    \label{tab:simpleExampleresult}
\end{table}

By calculating the utility, we can identify the top-\textit{k} itemsets as the most strongly associated patterns, including \{``Age\_Low'', ``Spending\_Medium''\} and so on. By analyzing these patterns, we can identify the association rules between the combinations of low, medium, high and categorical features. This helps users to interpret the results of the prediction model.

\begin{table}
\caption{Incorporation of the top-1 identified fuzzy patterns as a new column in the training set.}
    \centering
    \begin{tabular}{|c|c|c|c|c|}
    \hline
    \textbf{ID} & \textbf{Shop Location} & \textbf{Age} & \textbf{Spending} & \textbf{*HAFCP*} \\
    \hline
    A & N (1)& 25 & 5000 & 1 \\
    B & S (2)& 30 & 3000 & 0 \\
    C & N (1)& 28 & 4500 & 1 \\
    D & C (3)& 55 & 7000 & 0 \\
    E & N (1)& 60 & 1000 & 0 \\
    F & S (2)& 35 & 6500 & 0 \\
    G & N (1)& 40 & 5500 & 0 \\
    H & C (3)& 65 & 3500 & 0 \\
    I & S (2)& 23 & 4500 & 1 \\
    J & N (1)& 50 & 3000 & 0 \\
    \hline
    \end{tabular}
    \label{tab:simpleExamplefeatureeng}
\end{table}
At the end, Table \ref{tab:simpleExamplefeatureeng} shows the results for using HAFCP to introduce new patterns in an additional column. This table illustrates how patterns identified by HAFCP are integrated into the dataset to demonstrate their application in feature engineering or data augmentation. By treating the discovered HAFCP as a new binary feature in the training set and then retraining our model, we increase the likelihood for achieving better results. In this case, the top-1 HAFCP \{``Age\_Low'', ``Spending\_Medium''\} is used as an example and the customers $A$, $C$ and $I$ are marked as $1$ to strengthen the training dataset. However, this is only one approach and its effectiveness can vary based on different characteristics of datasets, different data may provide different levels of benefit.

\section{Case Study and Discussion}

In this section, we first introduce the datasets and then explain the analytical framework tailored to the \cite{AMIN2023110103,customer-churn-prediction-2020,app12189355,misc_iranian_churn_dataset_563} used in this study and discuss the evaluation measures. At the end of this section, several experiments are then conducted to evaluate the performance of the proposed approach and the compared models.
%
In this study, AUC, precision, accuracy, F1 score, and recall rate will be employed as metrics for measuring model prediction performance \cite{hossin2015review}.
%

We use four publicly available datasets \cite{AMIN2023110103,customer-churn-prediction-2020,app12189355,misc_iranian_churn_dataset_563} from the telecommunications sector, and one dataset from the banking sector specifically targeting credit card customers\footnote{Link: \url{https://www.kaggle.com/code/kmalit/bank-customer-churn-prediction/notebook}}. The target class denoting the churned customer is labeled as "True" or "Exit" in these datasets. The datasets from telecom industry provide information about customers' telephony usage, including call durations, charges, frequency of different types of calls, customers' geographical location, account tenure, subscription plans, and interactions with customer service. The dataset from the banking industry mainly consists of customer usage information, such as personal details including nationality, gender, tenure, and credit card usage records, etc. Dataset 1 contains 3,333 entries with 15 numerical attributes and 5 categorical attributes. Dataset 2 contains 5,000 entries with the same attributes as the first dataset. Dataset 3 contains 3,150 instances with 13 attributes. Dataset 4 contains 51,047 labeled data entries with 57 attributes. Dataset 5 contains 10,000 labeled data entries with 15 attributes.

We first used the XGBoost \cite{chen2016xgboost} ML model with the parameter settings as: $max\_depth=6$, $learning\_rate=0.3$, and $n\_estimators=100$. The performance metrics of the baseline model for datasets 1-5 are shown in Table \ref{tab:datasets_performance_enhancement}.
%
We then extract numerical columns from the dataset. For datasets 1 and 2, four features relating to the ``Total Charge'' are filtered out directly to save computational effort, as the importance of the feature is zero. In addition, features with low importance and several categories, namely ``STATE'' and ``AREA'', are eliminated. After performing normality tests on these columns, Gaussian membership functions \cite{jang1997neuro} (for columns with normal distribution) and triangular membership functions \cite{jang1997neuro} (for columns without normal distribution) are applied, and then the maximal cardinality is selected to categorize each column based on the predefined membership functions. This process of fuzzification is followed by one-hot encoding of the fuzzy classification columns. While the same logic and methodology is applied to datasets 3, 4 and 5, the specifics of the data differ, so the membership functions must be adjusted to accurately reflect the unique distributions in these additional datasets.

From the above results, we have already obtained a dataset that combines categorical data with a fuzzy classification dataset. Now we have a table of frequent itemsets and unit profits. The results of the top-\textit{k} HUIM on the datasets calculated Algorithm \ref{algo1} by show the mining results of datasets that consider the maximum cardinality as a binary value for fuzzy columns. In this context, the suffix ``\_L'' after a column name stands for the linguistic expression of ``low'' after fuzzy classification, ``\_M'' stands for ``medium'' and ``\_H'' for ``high''. For instance, in dataset 1, the top-1 HAFCP found was \{'TTL DAY MIN\_L', 'TTL INTL MIN\_M'\}.

Finally, we extract the association rules of the top-5 patterns to see if they can improve the performance of the model in these datasets. In this experiment, we tested by generating the binary column resulting from multiplying different features into the dataset and then training the dataset with the same parameter settings. By including the top-5 HAFCP separately, the models for both datasets underwent subtle but notable changes, as shown in Table \ref{tab:datasets_performance_enhancement}. It shows the evaluation results from two previous studies \cite{pamina2019effective, wu2021integrated} (one of which is the baseline) and compares them with the results obtained by including the top-5 HAFCP methods.

\begin{table}[!ht]
    \centering
    \caption{Comparison of performance metrics with adding top 1-5 highly associated fuzzy churn patterns for datasets 1-5. Results indicating improved performance compared to the XGBoost \cite{pamina2019effective} and RF \cite{wu2021integrated} as the baseline models.}
    \begin{tabular}{|c|l|c|c|c|c|c|c|c|c|}
    \hline
        \multicolumn{10}{|c|}{\textbf{Performance Metrics}} \\
    \hline
        \textbf{Dataset} & \textbf{Metrics} & \textbf{RF} & \textbf{XGBoost} & \textbf{Top-1} & \textbf{Top-2} & \textbf{Top-3} & \textbf{Top-4} & \textbf{Top-5} & \textbf{AVG} \\
    \hline
        \multirow{5}{*}{1} & AUC & 91.3\% & 91.2\% & \textbf{92.5\%} & \textbf{93.1\%} & \textbf{92.9\%} & \textbf{92.9\%} & \textbf{92.4\%} & \textbf{92.8\%} \\
         & Accuracy & 95.3\% & 95.7\% & 95.7\% & 95.7\% & \textbf{96.0\%} & 95.7\% & 95.7\% & \textbf{95.8\%} \\
         & Recall & 71.9\% & 76.2\% & 76.2\% & 76.2\% & \textbf{79.2\%} & 76.2\% & 76.2\% & \textbf{76.8\%} \\
         & Precision & 91.7\% & 93.9\% & 93.9\% & 93.9\% & 93.0\% & 93.9\% & 93.9\% & 93.7\% \\
         & F1 Score & 80.4\% & 84.2\% & 84.2\% & 84.2\% & \textbf{85.6\%} & 84.2\% & 84.2\% & \textbf{84.5\%} \\
    \hline
        \multirow{5}{*}{2} & AUC & 89.9\% & 90.6\% & 90.6\% & 90.5\% & \textbf{90.9\%} & 90.6\% & \textbf{90.9\%} & \textbf{90.7\%}\\
        & Accuracy & 94.5\% & 95.6\% & 95.6\% & 95.4\% & 95.6\%  & \textbf{96.0\%} & \textbf{95.7\%} & \textbf{95.7\%}\\
        & Recall & 78.9\% & 77.9\% & 77.9\% & 76.9\% & \textbf{79.0\%} & \textbf{80.0\%} & \textbf{79.0\%} & \textbf{78.6\%} \\
        & Precision & 84.1\% & 92.1\% & 92.1\% & 92.0\% & 91.1\% & \textbf{92.7\%} & 91.7\% & \textbf{92.0\%} \\
        & F1 Score & 81.5\% & 84.4\% & 84.4\% & 83.8\% & \textbf{84.6\%} & \textbf{86.0\%} & \textbf{84.9\%} & \textbf{84.7\%} \\
    \hline
    \multirow{5}{*}{3} & AUC & 97.6\% & 98.5\% & \textbf{98.6\%} & \textbf{98.6\%} & \textbf{98.6\%} & \textbf{98.6\%} & \textbf{98.6\%} & \textbf{98.6\%} \\
    & Accuracy & 94.3\% & 96.2\% & \textbf{96.7\%} & \textbf{96.7\%} & \textbf{96.7\%} & \textbf{96.7\%} & \textbf{96.7\%} & \textbf{96.7\%} \\
    & Recall & 85.5\% & 85.5\% & \textbf{86.7\%} & \textbf{86.7\%} & \textbf{86.7\%} & \textbf{86.7\%} & \textbf{86.7\%} & \textbf{86.7\%} \\
    & Precision & 82.5\% & 85.5\% & \textbf{87.8\%} & \textbf{87.8\%} & \textbf{87.8\%} & \textbf{87.8\%} & \textbf{87.8\%} & \textbf{87.8\%} \\
    & F1 Score & 83.9\% & 85.5\% & \textbf{87.2\%} & \textbf{87.2\%} & \textbf{87.2\%} & \textbf{87.2\%} & \textbf{87.2\%} & \textbf{87.2\%} \\
\hline
    \multirow{5}{*}{4} & AUC & 62.9\% & 65.0\% & \textbf{65.5\%} & \textbf{65.3\%} & 65.0\% & 65.0\% & \textbf{65.6\%} & \textbf{65.3\%} \\
    & Accuracy & 69.2\% & 64.0\% & 63.9\% & \textbf{64.1\%} & 63.7\% & 63.7\% & \textbf{64.2\%} & 63.9\% \\
    & Recall & 18.3\% & 51.6\% & \textbf{53.1\%} & \textbf{52.5\%} & \textbf{51.7\%} & \textbf{51.7\%} & \textbf{53.3\%} & \textbf{52.5\%} \\
    & Precision & 40.8\% & 39.4\% & \textbf{39.9\%} & \textbf{40.0\%} & \textbf{39.5\%} & \textbf{39.5\%} & \textbf{40.4\%} & \textbf{39.9\%} \\
    & F1 Score & 25.3\% & 44.7\% & \textbf{45.6\%} & \textbf{45.4\%} & \textbf{44.8\%} & \textbf{44.8\%} & \textbf{45.9\%} & \textbf{45.3\%} \\
\hline
    \multirow{5}{*}{5} & AUC & 80.5\% & 80.1\% & \textbf{88.9\%} & \textbf{89.9\%} & \textbf{88.9\%} & \textbf{88.9\%} & \textbf{91.2\%} & \textbf{89.4\%} \\
    & Accuracy & 79.1\% & 82.9\% & \textbf{91.3\%} & \textbf{91.3\%} & \textbf{91.3\%} & \textbf{91.3\%} & \textbf{93.4\%} & \textbf{91.73\%} \\
    & Recall & 63.4\% & 41.9\% & \textbf{63.6\%} & \textbf{63.6\%} & \textbf{63.6\%} & \textbf{63.6\%} & \textbf{71.5\%} & \textbf{65.2\%} \\
    & Precision & 47.5\% & 59.4\% & \textbf{88.9\%} & \textbf{88.9\%} & \textbf{88.9\%} & \textbf{88.9\%} & \textbf{93.7\%} & \textbf{90.0\%} \\
    & F1 Score & 54.3\% & 49.2\% & \textbf{74.2\%} & \textbf{74.2\%} & \textbf{74.2\%} & \textbf{74.2\%} & \textbf{81.1\%} & \textbf{75.6\%} \\
\hline
    \end{tabular}
    \label{tab:datasets_performance_enhancement}
\end{table}

\begin{figure}[!ht]
    \centering
    \begin{minipage}{0.47\textwidth}
        \includegraphics[width=\textwidth]{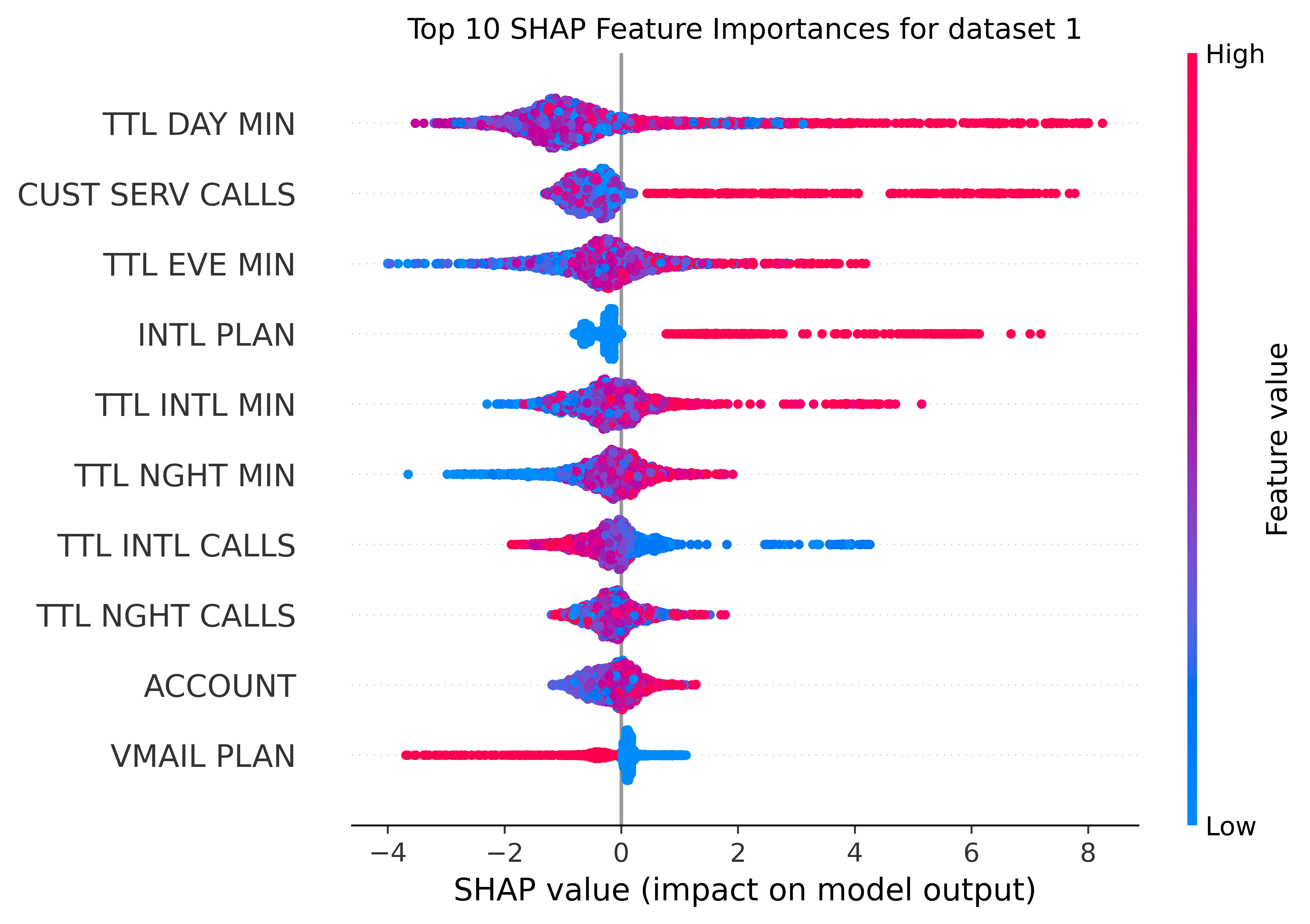}
    \end{minipage}
    \hfill
    \begin{minipage}{0.47\textwidth}
        \includegraphics[width=\textwidth]{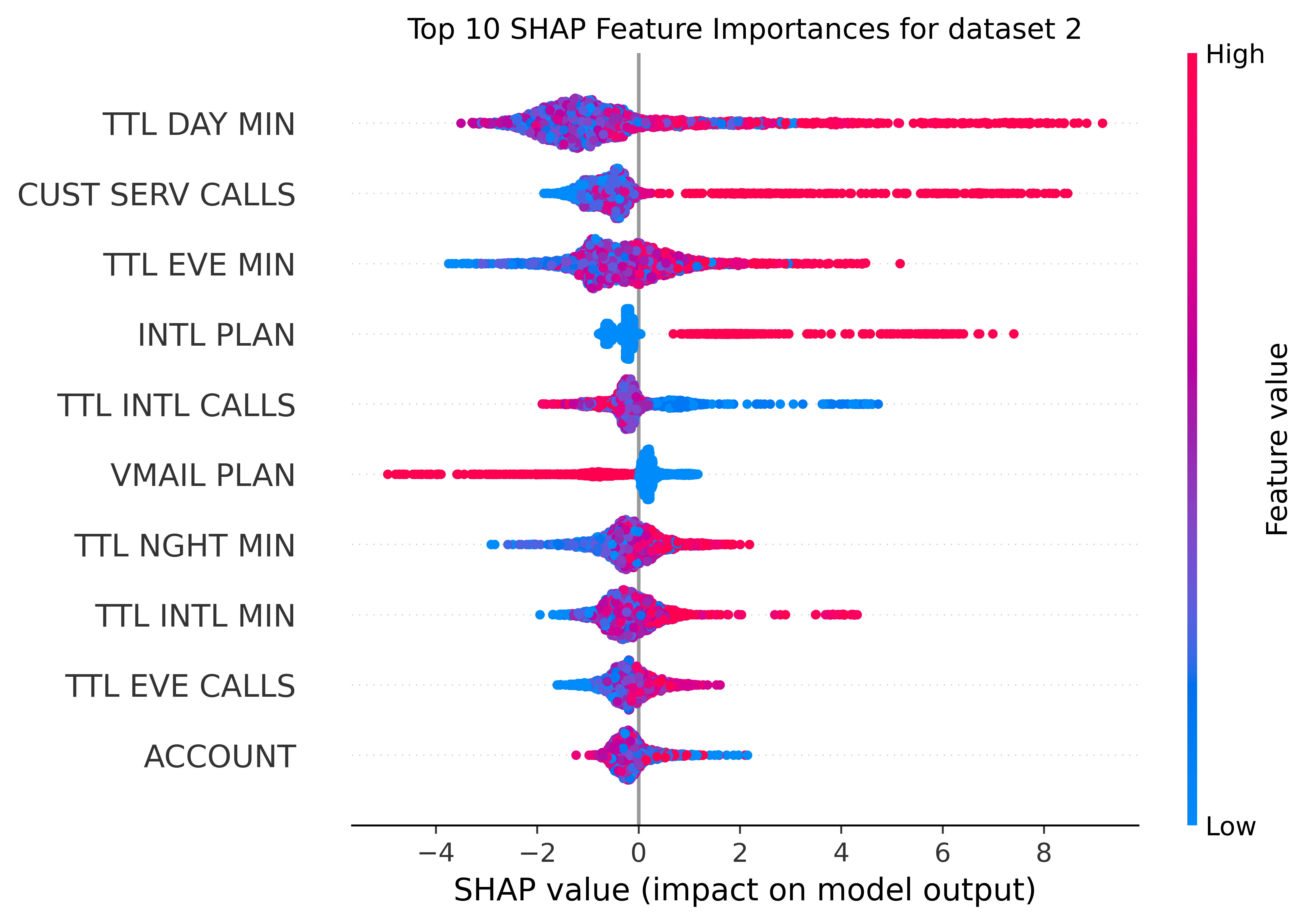}
    \end{minipage}

    \begin{minipage}{0.47\textwidth}
        \includegraphics[width=\textwidth]{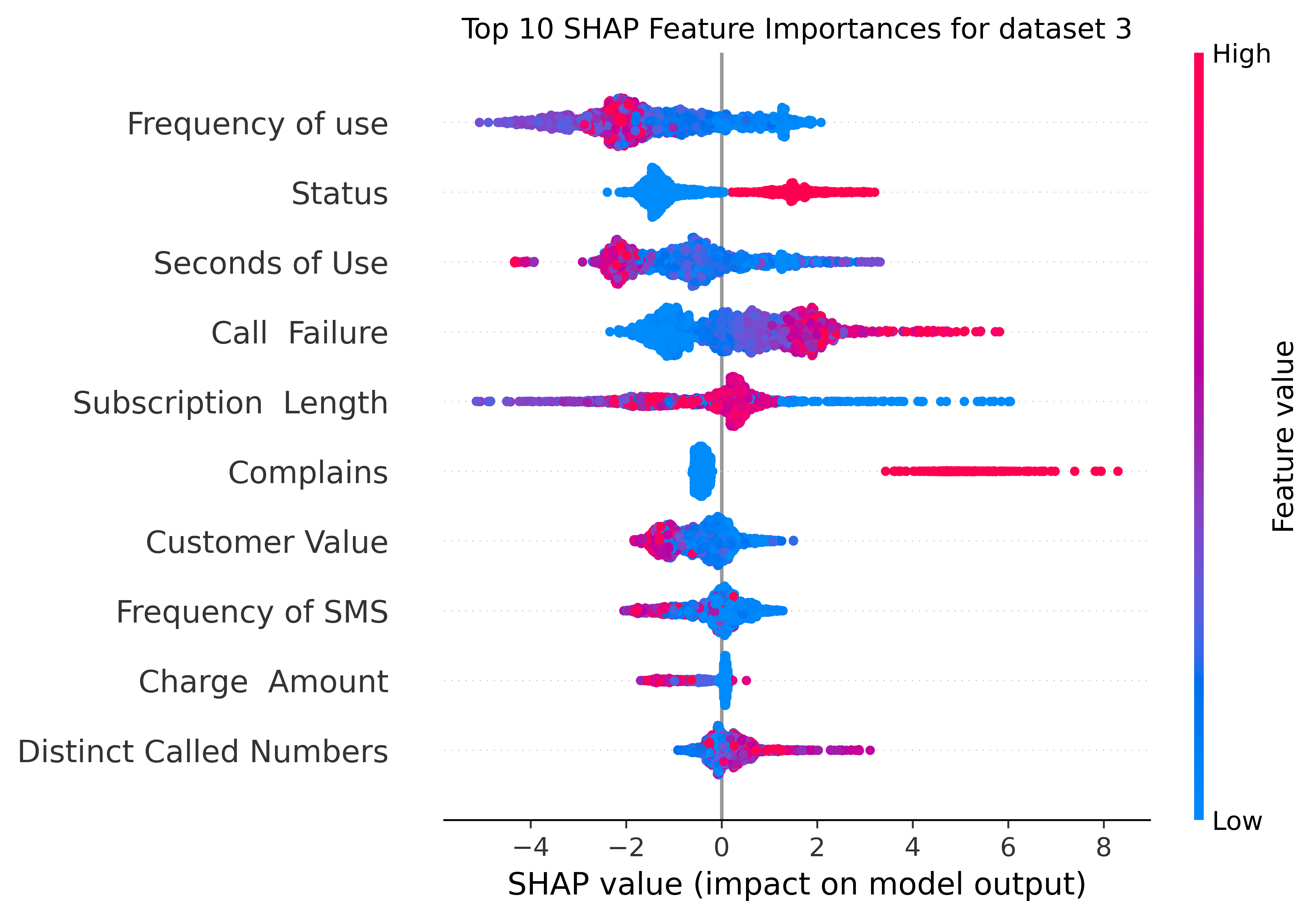}
    \end{minipage}
    \hfill
    \begin{minipage}{0.47\textwidth}
        \includegraphics[width=\textwidth]{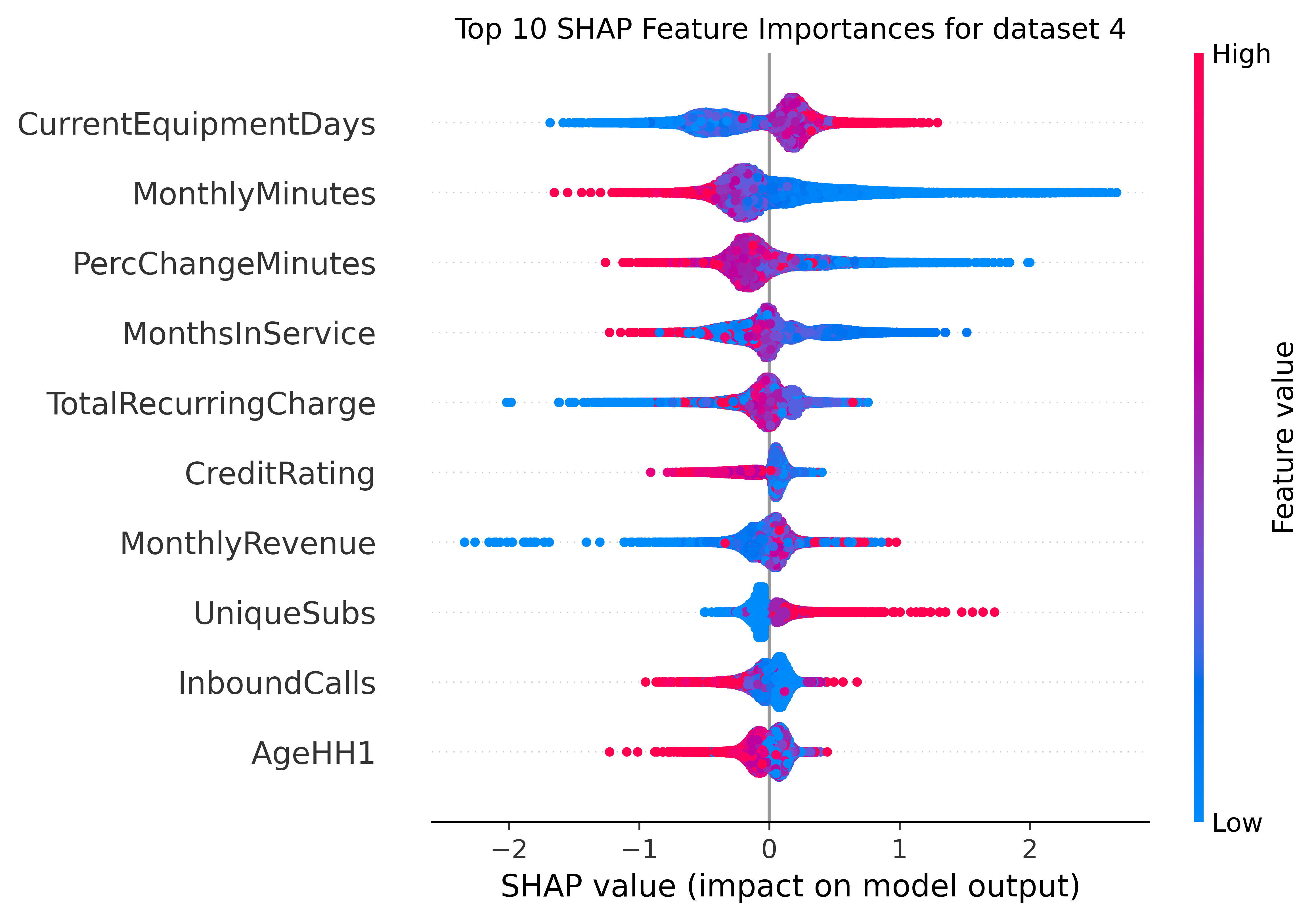}
    \end{minipage}
    \hfill
    \begin{minipage}{0.47\textwidth}
        \includegraphics[width=\textwidth]{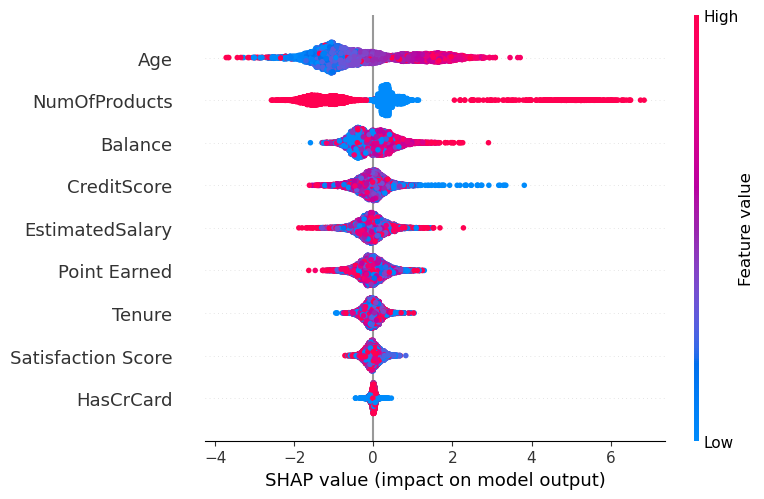}
    \end{minipage}
    \caption{Top-10 features of explainability in five datasets}\label{fig3}
\end{figure}

The extended evaluation over five datasets shows the robustness and adaptability of our model. Remarkably, a detailed observation reveals that Dataset 1 experienced a solid increase in AUC, especially noting a 1.5\% growth from the top-5 patterns. Additionally, the recall metric showed significant improvement with the top-3 patterns, indicating more accurate and positive predictions. Dataset 2 is modest but consistent enhancements in accuracy and precision, despite minimal fluctuations in AUC and recall, suggesting a positive effect of HAFCP inclusion. The high stability of performance metrics in Dataset 3 demonstrates the model's consistency. In Dataset 4, significant improvements in recall and precision were observed, even from lower starting points. With this approach, prediction performance is not only maintained but even improved in several datasets, confirming the integration of fuzzy churn patterns as a valuable strategy for customer churn prediction. Dataset 5 showed remarkable improvements across several metrics especially in AUC, which increased by over 10\% when the top-5 patterns were added. The accuracy also showed a significant enhancement, moving from 82.9\% in the XGBoost model to 91.73\% with the inclusion of fuzzy churn patterns. This dataset highlights the potential of HAFCP to enhance predictive performance, especially for scenarios where baseline models may struggle. The marked increase in recall and F1 further indicates that approach captures the underlying patterns leading to customer churn in more challenging datasets. For detailed code implementation example, please refer to the provided GitHub\footnote{Githhub: \url{https://github.com/dannyycwang/HAFCP-algo/tree/main}}

Regarding the explainability of the prediction model, the meaning of individual features was extended to the association rules of multiple features under different distributions by HAFCP, which are determined by the meaning of the features. This can help us to confirm the correlation or causality between features with high association. Fig. \ref{fig3} shows a top-10 single feature visualization of the explainable models for five datasets. The associative patterns among the key variables influencing model performance. When combined with the visual insights from Fig. \ref{fig3}, these patterns facilitate a more comprehensive and intuitive interpretation of the model's behavior.
 
While these patterns seem to improve the performance of the model to a great extent, they can also have certain negative effects. So when incorporating new patterns to improve the performance of the model, it is important to carefully evaluate the pros and cons of the different metrics. Obtaining expert opinions on this topic would be of great benefit. In addition, it is essential to identify potential areas for refinement of this methodology. Categorical variables often play an important role in ML models. However, since high-utility mining focuses on frequency, certain categorical variables, even those with significant feature importance, could be sidelined if they are not sufficiently frequent. An example from this dataset is the variable ``INTL PLAN''. Future efforts should address more efficient strategies for handling categorical variables in this framework.

\section{Conclusion}
To achieve a better understanding of customer churn models and uncover the associative patterns between different feature distributions, this paper presents a method for identifying strongly associated fuzzy churn patterns in tabular datasets with numerical values and provides a comprehensive guide and examples for its application. The method effectively utilizes fuzzy set theory and enables the transformation of numerical data into frequent item sets in a defensible manner. In addition, the associated patterns can provide insights for the interpretation of prediction models. As part of a case study, we used five publicly available customer churn datasets to validate the feasibility and pilot runs of our approach and its potential to improve model performance.

\bibliographystyle{splncs04}
\bibliography{ref.bib}

\begin{thebibliography}{10}
\providecommand{\url}[1]{\texttt{#1}}
\providecommand{\urlprefix}{URL }
\providecommand{\doi}[1]{https://doi.org/#1}

\bibitem{altmann2010permutation}
Altmann, A., Tolo{\c{s}}i, L., Sander, O., Lengauer, T.: Permutation importance: a corrected feature importance measure. Bioinformatics  \textbf{26}(10),  1340--1347 (2010)

\bibitem{AMIN2023110103}
Amin, A., Adnan, A., Anwar, S.: An adaptive learning approach for customer churn prediction in the telecommunication industry using evolutionary computation and naïve bayes. Applied Soft Computing  \textbf{137} (2023)

\bibitem{chen2016xgboost}
Chen, T., Guestrin, C.: Xgboost: A scalable tree boosting system. ACM SIGKDD International Conference on Knowledge Discovery and Data Mining pp. 785--794 (2016)

\bibitem{customer-churn-prediction-2020}
Diamantaras, K.: Customer churn prediction 2020 (2020), \url{https://kaggle.com/competitions/customer-churn-prediction-2020}

\bibitem{fournier2019survey}
Fournier-Viger, P., Chun-Wei~Lin, J., Truong-Chi, T., Nkambou, R.: A survey of high utility itemset mining. High-Utility Pattern Mining: Theory, Algorithms and Applications pp. 1--45 (2019)

\bibitem{gan2021explainable}
Gan, W., Du, Z., Ding, W., Zhang, C., Chao, H.C.: Explainable fuzzy utility mining on sequences. IEEE Transactions on Fuzzy Systems  \textbf{29}(12),  3620--3634 (2021)

\bibitem{giacometti2019mining}
Giacometti, A., Markhoff, B., Soulet, A.: Mining significant maximum cardinalities in knowledge bases. In: International Semantic Web Conference. pp. 182--199. Springer (2019)

\bibitem{guliyev2021customer}
Guliyev, H., Yerdelen~Tato{\u{g}}lu, F.: Customer churn analysis in banking sector: Evidence from explainable machine learning model. Journal of Applied Microeconometrics  \textbf{1}(2) (2021)

\bibitem{hancock2020survey}
Hancock, J.T., Khoshgoftaar, T.M.: Survey on categorical data for neural networks. Journal of Big Data  \textbf{7}(1),  1--41 (2020)

\bibitem{hong2003fuzzy}
Hong, T.P., Lin, K.Y., Wang, S.L.: Fuzzy data mining for interesting generalized association rules. Fuzzy sets and systems  \textbf{138}(2),  255--269 (2003)

\bibitem{hossin2015review}
Hossin, M., M.N, S.: A review on evaluation metrics for data classification evaluations. International Journal of Data Mining and Knowledge Management Process  \textbf{5},  1--11 (2015)

\bibitem{jang1997neuro}
Jang, J.S.R., Sun, C.T., Mizutani, E.: Neuro-fuzzy and soft computing-a computational approach to learning and machine intelligence [book review]. IEEE Transactions on Automatic Control  \textbf{42}(10),  1482--1484 (1997)

\bibitem{LI2022101845}
Li, Z.: Extracting spatial effects from machine learning model using local interpretation method: An example of shap and xgboost. Computers, Environment and Urban Systems  \textbf{96},  101845 (2022)

\bibitem{liu2012mining}
Liu, M., Qu, J.: Mining high utility itemsets without candidate generation. In: Proceedings of the 21st ACM International Conference on Information and Knowledge Management. pp. 55--64 (2012)

\bibitem{app12189355}
Liu, R., Ali, S., Bilal, S.F., Sakhawat, Z., Imran, A., Almuhaimeed, A., Alzahrani, A., Sun, G.: An intelligent hybrid scheme for customer churn prediction integrating clustering and classification algorithms. Applied Sciences  \textbf{12}(18) (2022)

\bibitem{lundberg2017unified}
Lundberg, S.M., Lee, S.I.: A unified approach to interpreting model predictions. Advances in Neural Information Processing Systems  \textbf{30} (2017)

\bibitem{pamina2019effective}
Pamina, J., Raja, B., SathyaBama, S., Sruthi, M., VJ, A., et~al.: An effective classifier for predicting churn in telecommunication. Journal of Advanced Research in Dynamical \& Control Systems  \textbf{11} (2019)

\bibitem{peng2022research}
Peng, K., Peng, Y.: Research on telecom customer churn prediction based on ga-xgboost and shap. Journal of Computer and Communications  \textbf{10}(11),  107--120 (2022)

\bibitem{razali2011power}
Razali, N.M., Wah, Y.B., et~al.: Power comparisons of shapiro-wilk, kolmogorov-smirnov, lilliefors and anderson-darling tests. Journal of Statistical Modeling and Analytics  \textbf{2}(1),  21--33 (2011)

\bibitem{misc_iranian_churn_dataset_563}
Repository, U.M.L.: {Iranian Churn Dataset}. UCI Machine Learning Repository (2020)

\bibitem{saeed2023explainable}
Saeed, W., Omlin, C.: Explainable ai (xai): A systematic meta-survey of current challenges and future opportunities. Knowledge-Based Systems  \textbf{263},  110273 (2023)

\bibitem{shakerin2019induction}
Shakerin, F., Gupta, G.: Induction of non-monotonic rules from statistical learning models using high-utility itemset mining. ArXiv Preprint ArXiv:1905.11226  (2019)

\bibitem{shapiro1965analysis}
Shapiro, S.S., Wilk, M.B.: An analysis of variance test for normality (complete samples). Biometrika  \textbf{52}(3-4),  591--611 (1965)

\bibitem{sina2022model}
Sina~Mirabdolbaghi, S.M., Amiri, B.: Model optimization analysis of customer churn prediction using machine learning algorithms with focus on feature reductions. Discrete Dynamics in Nature and Society  \textbf{2022}(1),  5134356 (2022)

\bibitem{tseng2015efficient}
Tseng, V.S., Wu, C.W., Fournier-Viger, P., Philip, S.Y.: Efficient algorithms for mining top-k high utility itemsets. IEEE Transactions on Knowledge and Data Engineering  \textbf{28}(1),  54--67 (2015)

\bibitem{5277408}
Wang, C.M., Chen, S.H., Huang, Y.F.: A fuzzy approach for mining high utility quantitative itemsets. In: IEEE International Conference on Fuzzy Systems. pp. 1909--1913 (2009). \doi{10.1109/FUZZY.2009.5277408}

\bibitem{wang2023}
Wang, D.Y., Jordanger, L.A., Lin, J.C.W.: Explainability of leverage points exploration for customer churn prediction. In: IEEE International Conference on Big Data. pp. 5997--6004 (2023)

\bibitem{wu2021integrated}
Wu, S., Yau, W.C., Ong, T.S., Chong, S.C.: Integrated churn prediction and customer segmentation framework for telco business. IEEE Access  \textbf{9},  62118--62136 (2021)

\bibitem{zadeh1978fuzzy}
Zadeh, L.A.: Fuzzy sets as a basis for a theory of possibility. Fuzzy sets and systems  \textbf{1}(1),  3--28 (1978)

\bibitem{systems10060258}
Zhang, Y., Qin, C.: A gaussian-shaped fuzzy inference system for multi-source fuzzy data. Systems  \textbf{10}(6) (2022)

\end{thebibliography}
\end{document}


\title{Appendix: Interpretable Customer Churn Prediction}
%
%
%
%
%
\maketitle              

\appendix
\section{A Simple Example}\label{subsec:simpleexample}
In this subsection, we will demonstrate how to calculate and identify HAFCP using a simple example. Since the data is very simple, all values related to the algorithms are simulated so that the reader can quickly grasp the concept. Let us assume a customer churn training dataset that contains the identities of the customers. For categorical variables, it contains the locations where customers shop ($South = S$, $Central = C$ and $North = N$). For numerical variables, it contains the age of the customer and their total spend. In addition, there is a label indicating whether the customer has churned, marked as $0=``Non-Churn''$ or $1=``Churn"$. This is shown in Table \ref{tab:simpleExample}. Following the steps of the analysis framework from the previous sub-session, a label encoding of the dataset must first be carried out (e.g., defining the shop location $N=1$, $S=2$, $C=3$). Then we create an ML prediction model. After training the ML model, we obtained the following feature importance: purchase location: $0.2$, age: $0.5$, purchase amount: $0.3$. The feature importance here is the value of profit used in high utility mining.
\begin{table}[ht]
    \caption{A simple customer churn dataset.}
    \centering
    \begin{tabular}{|c|c|c|c|c|}
    \hline
    \textbf{ID} & \textbf{Shop Location} & \textbf{Age} & \textbf{Spending} & \textbf{Churn (Given)} \\
    \hline
    A & N (1)& 25 & 5000 & 1 \\
    B & S (2)& 30 & 3000 & 0 \\
    C & N (1)& 28 & 4500 & 1 \\
    D & C (3)& 55 & 7000 & 0 \\
    E & N (1)& 60 & 1000 & 1 \\
    F & S (2)& 35 & 6500 & 0 \\
    G & N (1)& 40 & 5500 & 1 \\
    H & C (3)& 65 & 3500 & 0 \\
    I & S (2)& 23 & 4500 & 1 \\
    J & N (1)& 50 & 3000 & 1 \\
    \hline
    \end{tabular}
    \label{tab:simpleExample}
\end{table}

Next, it needs to categorize our numerical variables into easily interpretable linguistic terms such as low, medium and high using fuzzy set theory based on membership functions. First, we need to confirm the distribution of the numerical data columns before we can decide whether to use a Gaussian membership function or a triangular membership function. So for the two columns $``Age''$ and $``Spending''$ we first need to check if they follow a normal distribution. In this example, we used the Shapiro-Wilk test \cite{razali2011power}, which is suitable for small samples of less than $50$. The results show that both columns are consistent with the null hypothesis. They are therefore considered to be normally distributed. Subsequently, the two normally distributed columns are divided into three different categories using the Gaussian fuzzy membership function \cite{jang1997neuro}: low, medium and high, which enables a more differentiated analysis and interpretation. The results according to the fuzzy-set theory and classification can be found in Table \ref{tab:simpleExamplefuzzy}.

\begin{table}[ht]
\caption{A simple customer churn dataset after transformation by the fuzzy-set theory.}
\centering
\begin{tabular}{|c|c|c|c|c|}
\hline
\textbf{ID} & \textbf{Shop Location} & \textbf{Age\_fuzzy} & \textbf{Spending\_fuzzy} & \textbf{Churn (Given)} \\
\hline
A & N (1) & Low  (0.97)& Medium (0.97) & 1 \\
B & S (2) & Low (0.99)& Low (0.99)& 0 \\
C & N (1) & Low (0.99)& Medium (1)& 1 \\
D & C (3) & Medium (0.99)& High (0.66)& 0 \\
E & N (1) & High (0.92)& Low (0.49)& 1 \\
F & S (2) & Medium (0.98)& High (0.82)& 0 \\
G & N (1) & Medium (0.98)& Medium (0.99)& 1 \\
H & C (3) & High (0.76)& Low (0.97)& 0 \\
I & S (2) & Low (0.93)& Medium (1)& 1 \\
J & N (1) & High (0.97)& Low (0.99)& 1 \\
\hline
\end{tabular}
\label{tab:simpleExamplefuzzy}
\end{table}

Since the problem we defined focuses on the utility patterns of churn, we first need to filter out data instances where $``Churn'' = 0$. As shown in Table \ref{tab:simpleExamplelabelencod} by simply performing One-hot encoding, we can convert our dataset into frequent itemsets.

\begin{table}[ht]
\caption{Generation of frequent itemsets from customer churn data, where SL is labeled as shop location, L as low, M as medium and H as high.}
\centering
\small 
\begin{tabular}{|c|c|c|c|c|c|c|c|c|c|c|}
\hline
\textbf{ID} & \textbf{SL\_C} & \textbf{SL\_N} & \textbf{SL\_S} & \textbf{Age\_L} & \textbf{Age\_M} & \textbf{Age\_H} & \textbf{Spend\_L} & \textbf{Spend\_M} & \textbf{Spend\_H} \\
\hline
A & 0 & 1 & 0 & 1 (0.97) & 0 & 0 & 0 & 1 (0.97) & 0 \\
C & 0 & 1 & 0 & 1 (0.99) & 0 & 0 & 0 & 1 (1) & 0 \\
E & 0 & 1 & 0 & 0 & 0 & 1 (0.92)& 1 (0.49)& 0 & 0 \\
G & 0 & 1 & 0 & 0 & 1 (0.98)& 0 & 0 & 1 (0.99)& 0 \\
I & 0 & 0 & 1 & 1 (0.93)& 0 & 0 & 0 & 1 (1)& 0 \\
J & 0 & 1 & 0 & 0 & 0 & 1 (0.97) & 1 (0.99)& 0 & 0 \\
\hline
\end{tabular}
\label{tab:simpleExamplelabelencod}
\end{table}

With the transactions of the frequent itemsets and the profit table, HUIM can be applied in this case to determine the associated fuzzy feature combinations with the maximum total utility. In contrast to setting a minimum threshold as in traditional HUIM methods, we use a top-k approach that allows us to identify the patterns with the highest utility in descending order. To save computational power and memory, rows consisting exclusively of zeros can be eliminated directly, since multiplication by the utility always results in zero. Therefore, the columns $``SL\_M''$ and $``Spending\_H''$ should be omitted before we proceed with the mining calculations. In this case, the utility of a frequent itemset is calculated by summing the product of the presence of each item and the corresponding profit from a given profit table, and searching for the top-k HUIs by iteratively expanding the itemsets and evaluating their utility. After each iteration, only the top-k itemsets with the highest utility are retained for further expansion. This process is repeated until no more itemsets can be expanded. Table \ref{tab:simpleExampleresult} shows the results for this example with slight variations noted between the two result sets for the utility value of the top-3 patterns.

\begin{table}[ht]
    \caption{Top-5 highly associated fuzzy churn patterns using binary value.}
    \centering
    \begin{tabular}{|c|c|c|}
    \hline
        \textbf{Rank} & \textbf{Highly Associated Fuzzy Churn Patterns} & \textbf{Utility} \\
        \hline
        1 & \{``Age\_Low", ``Spending\_Medium"\} & 2.0 \\
        2 & \{``Age\_Low", ``SL\_N", ``Spending\_Medium"\} & 2.0 \\
        3 & \{``SL\_N", ``Spending\_Medium"\} & 2.0 \\
        4 & \{``Age\_Low", ``SL\_N"\} & 1.0 \\
        5 & \{``Age\_Medium", ``SL\_N"\} & 1.0 \\
        \hline
    \end{tabular}
    \label{tab:simpleExampleresult}
\end{table}


By calculating the utility, we can identify the k most strongly associated patterns, including \{``Age\_Low'', ``Spending\_Medium''\} and so on. By analyzing these patterns, we can identify the association rules between the combinations of low, medium, high and categorical features. This helps users to interpret the prediction models effectively.

\begin{table}
\caption{Incorporation of the identified fuzzy patterns as a new column in the training set.}
    \centering
    \begin{tabular}{|c|c|c|c|c|}
    \hline
    \textbf{ID} & \textbf{Shop Location} & \textbf{Age} & \textbf{Spending} & \textbf{*HAFCP*} \\
    \hline
    A & N (1)& 25 & 5000 & 1 \\
    B & S (2)& 30 & 3000 & 0 \\
    C & N (1)& 28 & 4500 & 1 \\
    D & C (3)& 55 & 7000 & 0 \\
    E & N (1)& 60 & 1000 & 0 \\
    F & S (2)& 35 & 6500 & 0 \\
    G & N (1)& 40 & 5500 & 0 \\
    H & C (3)& 65 & 3500 & 0 \\
    I & S (2)& 23 & 4500 & 1 \\
    J & N (1)& 50 & 3000 & 0 \\
    \hline
    \end{tabular}
    \label{tab:simpleExamplefeatureeng}
\end{table}

At the end, Table \ref{tab:simpleExamplefeatureeng} shows a method for using HAFCP to introduce new patterns in an additional column. This table illustrates how patterns identified by HAFCP are integrated into the dataset to demonstrate their application in feature engineering or data augmentation. By treating the discovered HAFCP as a new binary feature in the training set and then retraining our model, we increase the likelihood of achieving better results. In this case, the top 1 HAFCP \{``Age\_Low'', ``Spending\_Medium''\} is used as an example and the customers $A$, $C$ and $I$ are marked as $1$ to strengthen the training dataset. However, this is only one approach and its effectiveness can vary from dataset to dataset. Different methods may provide different levels of benefit.

\section{Feature Explanation for Two Datasets}
The meanings of each feature in dataset 1 and 2 are elaborated upon in Table \ref{tab:features_information1,2} for example.
	\begin{table}[!ht]
		\centering
		\caption{Features Information of the Dataset 1 and 2}
		\label{tab:features_information1,2}
		\begin{tabular}{|l|l|}
			\hline
			\multicolumn{1}{|c|}{\textbf{Abbreviation}} & \multicolumn{1}{c|}{\textbf{Description}} \\
			\hline
			\textit{STATE} & 2-letter code of the US state\\
			\textit{ACCOUNT} & Customer's current provider tenure (in months)\\
			\textit{AREA CODE} & Area code of $3$ digits\\
			\textit{INTL PLAN} & The customer with international plan\\
			\textit{VMAIL PLAN} & The customer with voice mail plan\\
			\textit{VMAIL MSG} & Amount of voice-mail messages\\
			\textit{TTL DAY MIN} & Total minutes of day calls\\
			\textit{TTL DAY CALLS} & Amount of day calls\\
			\textit{TTL DAY CHARGE} & Total charge of day calls \\
			\textit{TTL EVE MINS} & Total minutes of evening calls\\
			\textit{TTL EVE CALLS} & Amount of evening calls\\
			\textit{TTL EVE CHARGE} & Total charge of evening calls\\
			\textit{TTL NIGHT MINS} & Total minutes of night calls\\
			\textit{TTL NIGHT CALLS} & Amount of night calls\\
			\textit{TTL NIGHT CHARGE} & Total charge of night calls\\
			\textit{TTL INTL MINS} & Total minutes of international calls\\
			\textit{TTL INTL CALLS} & Amount of international calls\\
			\textit{TTL INTL CHARGE} & Total charge of international calls\\
			\textit{CUST SERV CALLS} & Amount of calls to customer service\\
			\hline
		\end{tabular}
	\end{table}

\bibliographystyle{splncs04}
\bibliography{ref.bib}